# Actionable Approaches to Promote Ethical AI in Libraries

**Helen Bubinger**
Humboldt-Universität zu Berlin, Germany
*helen.bubinger@student.hu-berlin.de*

**Jesse David Dinneen**
Humboldt-Universität zu Berlin, Germany
*jesse.dinneen@hu-berlin.de*

## ABSTRACT

The widespread use of artificial intelligence (AI) in many domains has revealed numerous ethical issues from data and design to deployment. In response, countless broad principles and guidelines for ethical AI have been published, and following those, specific approaches have been proposed for how to encourage ethical outcomes of AI. Meanwhile, library and information services too are seeing an increase in the use of AI-powered and machine learning-powered information systems, but no practical guidance currently exists for libraries to plan for, evaluate, or audit the ethics of intended or deployed AI. We therefore report on several promising approaches for promoting ethical AI that can be adapted from other contexts to AI-powered information services and in different stages of the software lifecycle.

## INTRODUCTION AND METHOD

As in most domains of the information society, artificial intelligence (AI) has begun entering libraries, for example in tools for collection and catalogue analysis, systems that recommend items to patrons, and robotic greeters (Feng, 2021; Massis, 2018). However, such systems must be developed, implemented, and maintained carefully to minimise the numerous ethical issues they entail, small and large, from data and design to deployment, including issues with bias, fairness, accountability, transparency, responsibility, environmental impact, and more (Bender *et al.*, 2021; Slavkovik, 2020). Scholarly and public responses to address such issues have ranged from general principles and broad guidelines to specific recommendations and even development-to-deployment auditing approaches. Such approaches cannot guarantee ethical outcomes, but rather aim to increase their chances (i.e. promote them). Yet to our knowledge, no analogous guidance exists for libraries, so it is unclear how a library considering implementing AI, or already using it, might act on the above concerns to, for example, audit the ethics of their systems, or identify and communicate their ethical requirements to software vendors. And such action may become not only ethically but legally necessary, for example if public institutions are required to show their AI systems' ethics have been audited or have certification (Cihon *et al.*, 2021; Naudé & Dimitri, 2021).

As part of a systematic review of AI research in other domains we collected over two hundred publications about AI ethics and closely related topics (e.g. data ethics). For the present study we identified among those publications fifty-one that, at face value, contain actionable content for promoting ethical outcomes of AI, such as usable guidelines, proposed methods like auditing frameworks, and lessons learned. Such publications come primarily from the fields of computer science, philosophy, and applied (e.g. medical or business) ethics, whereas none are from library and information science (LIS) research. Though none of the publications were explicitly about libraries or related institutions, we read and discussed each more deeply to draw out any implicit relevance, potential applicability to, or usefulness for libraries. Below we discuss the few examples we find the most notable, grouped by theme and ordered according to when in the software lifecycle they could be used (Lauesen, 2002).

## RESULTS AND DISCUSSION

*Consult AI ethics guidelines (before and during requirements analysis)* -- States, organisations, human rights defenders, and other groups have produced countless guidelines for ethical AI that may be relevant to a library given their geographic location, mission, patrons' demographics, or the kind of AI in question. One useful starting point for considering which guidelines may be applicable or useful to a library is the searchable inventory of guidelines maintained by *AlgorithmWatch* (https://inventory.algorithmwatch.org/), which includes guidelines for/by particular regions and domains. Further, there are systematic reviews and syntheses of many guidelines from different perspectives, which can help give a useful overview, understand their uses and limitations (Hagendorff, 2020), and quickly establish which are relevant to an institution based on (a) their concerns (Ryan & Stahl, 2020) or (b) their commitment to the different principles motivating most guidelines (e.g. transparency, justice and fairness, non-maleficence, responsibility, privacy; Jobin *et al.*, 2019). We particularly recommend libraries consider which guidelines are important to them as early as possible so that they can be used in software selection and communicated to the vendor during functional requirements specification.

*Address bias (before and during design)* -- Bias is a problem that plagues AI in several forms (Mehrabi *et al.*, 2019), and there are many technical and non-technical approaches to address bias before and during the design of AI systems. We recommend relevant staff review all kinds of bias via a broad and generalised review (e.g. Ntoutsi *et al.*, 2020), and talk to vendors about if and how they audit their systems and training data for bias. For example, libraries might request developers use an auditing framework and questionnaire designed to uncover bias and "facilitate communication with non-technical stakeholders" (Lee & Singh, 2021). To further identify and address possible *non-technical* bias, libraries could organise events that draw on the perspectives of trustees, IT staff, LIS researchers, and the AI system's developers; consensus-building methods like ethical foresight analysis are generally designed for larger scales of AI, but could be useful in a library-AI context as well (Floridi & Strait, 2020).

*Audit the systems, development process, and data (throughout the software lifecycle)* -- Auditing is an aspect of evaluating and monitoring AI systems once they are implemented to ensure they are performing in an ethical manner, for example by systematically considering myriad possible unintended consequences. For an overview of issues that auditing can identify and address, we recommend a recent report by Koshiyama *et al.* (2021). However, because issues of AI can stem from its design and be hard to identify until after deployment, several auditing strategies, called end-to-end approaches, have been proposed to be put in place from the moment of identifying the perceived need for a system, *through* its development, to deployment and ongoing monitoring.

Notably, the SMACTR framework (Raji *et al.*, 2020) provides a promising guide to particular auditing steps (Scoping, Mapping, Artefact Collection, Testing, Reflection, and post-audit) and the necessary documents (e.g. list of principles, cases, checklists, data sheets, risk analyses, reports...) to conduct a thorough end-to-end algorithmic audit. We suspect it is sufficiently generic and thorough to be applicable and useful in libraries, but this should be validated. Though less generic and containing some examples that will not translate, there is also promise in an approach developed for auditing applications of machine learning to health care (Char *et al.*, 2020), because it is a very practical high-level guide to systematically identifying issues, with illuminating examples. If these solutions are too onerous given the scale of some system or the resources available for an audit, the Glass-Box Approach (Tubella *et al.*, 2019) may be a better fit; it is essentially a thoughtful adaptation of careful requirements specification and system verification to AI systems, consisting of (a) translating ethical values into design requirements and then (b) observing the deployed AI to see if the requirements are being met. Finally, though the previous frameworks do not ignore the importance of data (e.g. training data) in ethical AI, data-intensive cases like applications of AI to research data management may call for a framework that specifically considers the key data-ethical aspects. One such framework maps ethical themes and particular issues to five phases of data projects: business understanding, data preparation, modelling, evaluation, and deployment (Saltz & Dewar, 2019). Approaches like these are so far only *proposed*, rather than validated, but appear *prima facie* promising starting points for auditing libraries' AI systems.

**CONCLUSION**

AI is, with all its potential for good and bad, entering libraries today. Libraries therefore have an opportunity to evaluate and minimise ethical issues of their AI-powered systems, and in that process can be leaders of ethical AI in the public sphere. Though LIS-specific research on how to do this does not currently exist, we suggest the above approaches, taken from non-library contexts but adaptable into several stages of the software lifecycle, could be useful starting points. However, we also recommend that LIS research starts to explore *library-specific strategies* for ensuring ethical outcomes of AI, especially end-to-end approaches and with particular attention to making them feasible for practitioners to implement (Morley *et al.*, 2019).

**REFERENCES**

Bender, E. M., Gebru, T., McMillan-Major, A., & Shmitchell, S. (2021, March). On the Dangers of Stochastic Parrots: Can Language Models Be Too Big? In *FAccT '21: Proceedings of the 2021 ACM Conference on Fairness, Accountability, and Transparency* (pp. 610-623).

Cihon, P., Kleinaltenkamp, M. J., Schuett, J., & Baum, S. D. (2021). AI Certification: Advancing Ethical Practice by Reducing Information Asymmetries. *IEEE Transactions on Technology and Society*.

Char D. S., Abràmoff, M. D. & Feudtner, C. (2020). Identifying Ethical Considerations for Machine Learning Healthcare Applications. *The American Journal of Bioethics, 20*(11), 7-17. https://doi.org/10.1080/15265161.2020.1819469

Feng, W. (2021). The Applications of Artificial Intelligence in Reading Promotion in Chinese University Libraries. *Diversity, Divergence, Dialogue - iConference 2021, 16th International Conference on Information,* Beijing, China, March 17–31. https://www.ideals.illinois.edu/handle/2142/109662

Floridi, L., & Strait, A. (2020). Ethical foresight analysis: What it is and why it is needed? *Minds and Machines, 30,* 77-97.

Hagendorff, T. (2020). The Ethics of AI Ethics: An Evaluation of Guidelines. *Minds & Machines, 30*(1), 99–120. https://doi.org/10.1007/s11023-020-09517-8

Jobin, A., Ienca, M., & Vayena, E. (2019). The global landscape of AI ethics guidelines. *Nature Machine Intelligence, 1*(9), 389-399.

Koshiyama, A., Kazim, E., Treleaven, P., Rai, P., Szpruch, L., Pavey, G., Ahamat, G., Leutner, F., Goebel, R., Knight, A., Adams, J., Hitrova, C., Barnett, J., Nachev, P., Barber, D., Chamorro-Premuzic, T., Klemmer, K., Gregorovic, M., Khan, S., & Lomas, E. (2021, January) Towards Algorithm Auditing: A Survey on Managing Legal, Ethical and Technological Risks of AI, ML and Associated Algorithms. Available at SSRN: https://ssrn.com/abstract=3778998

Lauesen, S. (2002). *Software requirements: styles and techniques*. Pearson Education.

Lee, M. S. A., & Singh, J. (2021). Risk Identification Questionnaire for Detecting Unintended Bias in the Machine Learning Development Lifecycle. In *AIES '21: Proceedings of the AAAI/ACM Conference on AI, Ethics, and Society*. http://dx.doi.org/10.2139/ssrn.3777093

Massis, B. (2018). Artificial intelligence arrives in the library. *Information and Learning Sciences, 119*(7/8) 456-459.

Mehrabi, N., Morstatter, F., Saxena, N., Lerman, K., & Galstyan, A. (2019). A survey on bias and fairness in machine learning. *arXiv* preprint arXiv:1908.09635.

Morley, J., Floridi, L., Kinsey, L., & Elhalal, A. (2020). From what to how: an initial review of publicly available AI ethics tools, methods and research to translate principles into practices. *Science and Engineering Ethics, 26*(4), 2141-2168.

Naudé, W., & Dimitri, N. (2021). Public Procurement and Innovation for Human-Centered Artificial Intelligence. IZA Discussion Paper No. 14021, available at SSRN: https://ssrn.com/abstract=3762891

Ntoutsi, E., Fafalios, P., Gadiraju, U., Iosifidis, V., Nejdl, W., Vidal, M. E., Ruggieri, S., Turini, F., Papadopoulos, S., Krasanakis, E., Kompatsiaris, I., Kinder-Kurlanda, K., Wagner, C., Karimi, F., Fernandez, M., Alani, H., Berendt, B., Kruegel, T., Heinze, C., ... & Staab, S. (2020). Bias in data-driven artificial intelligence systems—An introductory survey. *Wiley Interdisciplinary Reviews: Data Mining and Knowledge Discovery, 10*(3), e1356.

Raji, I. D., Smart, A., White, R. N., Mitchell, M., Gebru, T., Hutchinson, B., Smith-Loud, J., Theron, D., & Barnes, P. (2020, January). Closing the AI accountability gap: defining an end-to-end framework for internal algorithmic auditing. In *Proceedings of the 2020 Conference on Fairness, Accountability, and Transparency* (pp. 33-44).

Ryan, M., Stahl., B. C. (2020). Artificial intelligence ethics guidelines for developers and users: clarifying their content and normative implications. *Journal of Information, Communication and Ethics in Society*, *19*(1), 61-86. https://doi.org/10.1108/JICES-12-2019-0138

Saltz, J.S. & Dewar, N. (2019). Data science ethical considerations: a systematic literature review and proposed project framework. *Ethics of Information Technology 21*, 197–208. https://doi.org/10.1007/s10676-019-09502-5

Slavkovik, M. (2020, November). Teaching AI Ethics: Observations and Challenges. In *Norsk IKT-konferanse for forskning og utdanning* (No. 4).

Tubella, A. A., Theodorou, A., Dignum, V., & Dignum, F. (2019). Governance by Glass-Box: Implementing Transparent Moral Bounds for AI Behaviour. *arXiv* preprint arXiv:1905.04994v2.
ASIS&T Annual Meeting 2021          5          Papers